\documentclass[11pt,a4paper]{article}
\usepackage[hyperref]{acl2021}
\usepackage{times}
\usepackage{latexsym}

\usepackage{booktabs}
\usepackage{adjustbox}
\usepackage{textcomp}
\usepackage{multirow}

\usepackage{xcolor}

\definecolor{orange}{RGB}{255, 100, 0}
\definecolor{blue}{RGB}{0, 76, 153}
\definecolor{brown}{RGB}{113, 30, 0}

\usepackage{microtype}

\aclfinalcopy 

\title{Can a Transformer Pass the Wug Test? Tuning Copying Bias in Neural Morphological Inflection Models}

\author{Ling Liu \and Mans Hulden\\
  University of Colorado \\
  {\tt first.last@colorado.edu} \\}

\date{}

\begin{document}
\maketitle
\begin{abstract}
Deep learning sequence models have been successfully applied to the task of morphological inflection. The results of the SIGMORPHON shared tasks in the past several years indicate that such models can perform well, but only if the training data cover a good amount of different lemmata, or if the lemmata that are inflected at test time have also been seen in training, as has indeed been largely the case in these tasks.  Surprisingly, standard models such as the Transformer almost completely fail at generalizing inflection patterns when asked to inflect previously unseen lemmata---i.e. under ``wug test''-like circumstances. While established data augmentation techniques can be employed to alleviate this shortcoming by introducing a copying bias through hallucinating synthetic new word forms using the alphabet in the language at hand, we show that, to be more effective, the hallucination process needs to pay attention to substrings of syllable-like length rather than individual characters or stems. We report a significant performance improvement with our substring-based hallucination model over previous data hallucination methods when training and test data do not overlap in their lemmata.


\end{abstract}

\section{Introduction}
\label{sec:intro}

The Transformer model has delivered convincing results in many different tasks related to word-formation and analysis \cite{vylomova-etal-2020-sigmorphon}. Especially on inflection tasks, where an input lemma such as {\tt dog}, and input inflectional features such as \{{\tt N,PL}\}, are expected to produce an output such as {\tt dogs}, the model has shown to be particularly adept at generalizing patterns \cite{wu2020applying,liu-hulden-2020-analogy}.  However, we have discovered that this is only true if {\it some} variant of the input lemma to be inflected has been witnessed during training.  In a ``wug test'' \cite{berko1958} setting where a previously unseen lemma---like {\tt wug}---is to be inflected in some way, we find that the Transformer almost completely fails to generalize inflection patterns, despite abundant training data.  It has been noted earlier that neural sequence-to-sequence models are apt to perform poorly if they have been exposed to little training data and that autoencoding on hallucinated forms could be useful \cite{kann-schutze-2017-unlabeled}. Our starting point is the observation that the poor ``wug test'' performance is maintained even with abundant training data. 

\begin{figure}
    \centering
    \includegraphics[width=.9\linewidth]{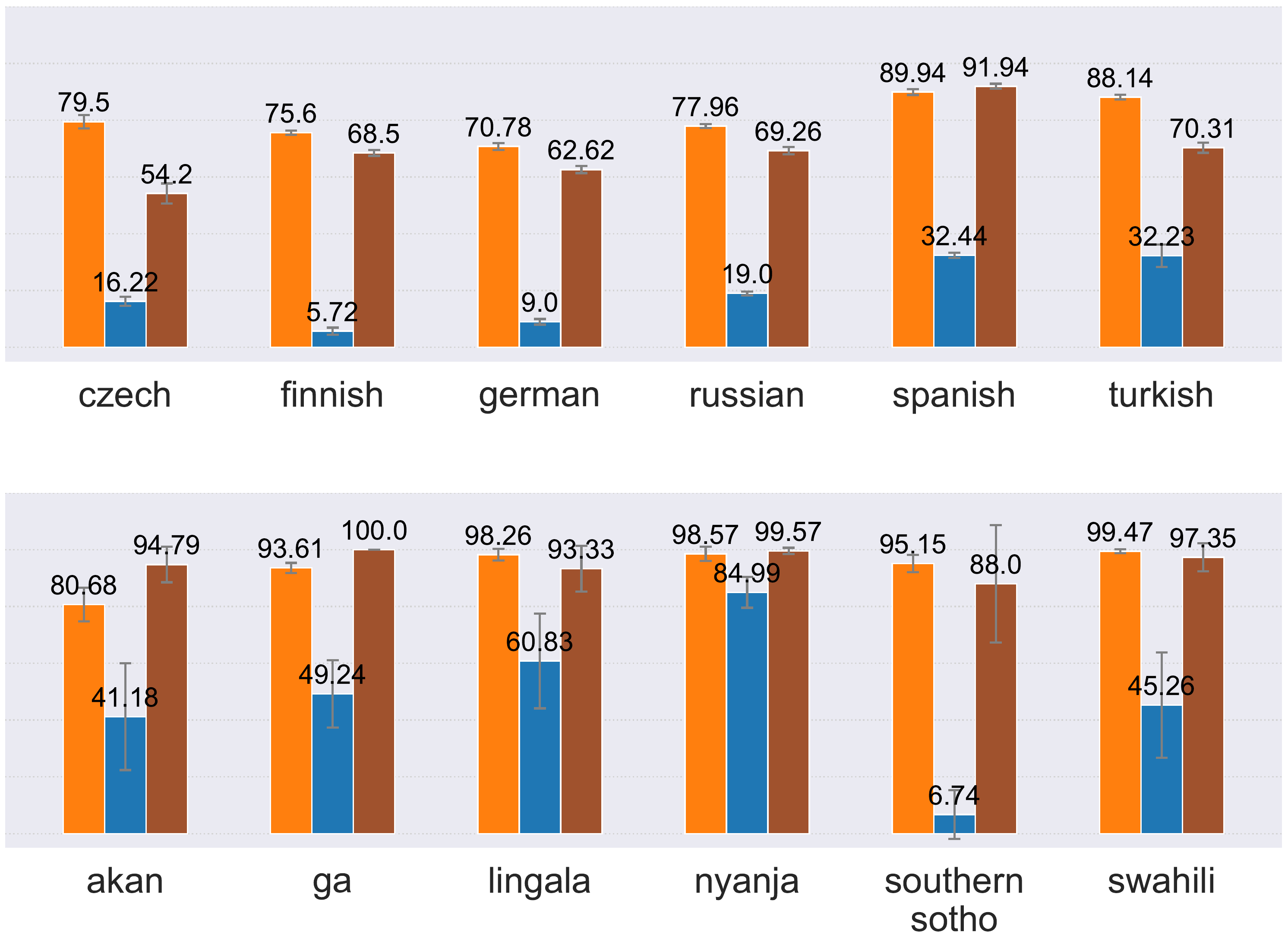}
    \caption{Transformer performance in the \textcolor{orange}{\textbf{common-practice}} setting (left), ``\textcolor{blue}{\textbf{wug test}}''-like setting (middle), and `\textcolor{brown}{\textbf{`wug test''-like setting with our best data hallucination}} method (right)}
    \label{fig:shared_tas_vs_full_table_vs_add_hall}
\end{figure}

In our study, we show three main results. (1) We demonstrate that, even if trained with relatively large amounts of data, a Transformer model of the kind that has been very successful at recent shared tasks largely fails to generalize inflection patterns if it has not been exposed during training to lemmata in the test set.  This is true even for datasets where all words inflect in the same way---i.e. there are no inflectional classes or allomorphs of morphemes, as is found in the low-resource Niger-Congo dataset used in SIGMORPHON 2020 shared task \cite{vylomova-etal-2020-sigmorphon}. (2) We also show that simply exposing the model to uninflected lemmata in the test set---without providing a single inflected form---allows the model to dramatically improve its performance when actually inflecting such lemmata. (3) Further, we investigate several strategies that avoid leveraging test set lemmata.  We show that when inducing a copy bias in the model by hallucinating new lemmata, or by hallucinating new inflected forms, the method of hallucination is much more effective if it is sensitive to substrings of syllable-like length rather than individual characters or stems. Our best models significantly improve upon earlier state-of-the-art data hallucination methods such as \newcite{silfverberg-etal-2017-data} and \newcite{anastasopoulos-neubig-2019-pushing}.



\section{Data}
\label{sec:data}

\paragraph{2018-languages} We use six languages from the CoNLL-SIGMORPHON 2018 shared task 1 medium setting, where each language has 1,000 \texttt{\small{(LEMMA, TARGET TAGS, TARGET FORM)}} triples for training \citep{cotterell-etal-2018-conll}. The six languages, Czech, Finnish, German, Russian, Spanish and Turkish, are selected to provide a diversified representation of language typology and morphological inflection challenges. Although there are only 1,000 triples in the training set, they cover a fair number of lemmata as each lemma appears only once or twice. In the original shared task data split, between 2\% and 27\% of the lemmata in the dev and test sets are also found in the training set. 

To prepare training data for the ``wug test''-like circumstance, we select the UniMorph \citep{kirov-etal-2018-unimorph} paradigms for the first 100 most frequent lexemes found in Wikipedia text, which are not included in the 2018 shared task 1 dev and test sets. The shared task dev and test sets are used for validation and evaluation without any change. The 100 full inflection tables give us over 1,000 (for Czech, German and Russian) or over 7,000 (for Finnish, Spanish and Turkish) training triples. 

\smallbreak
\noindent \textbf{Niger-Congo languages} \ \ \ In addition, we use six Niger-Congo languages from SIGMORPHON 2020 shared task 0 \citep{vylomova-etal-2020-sigmorphon}: Akan, Ga, Lingala, Nyanja, Southern Sotho and Swahili. These languages are low-resource, but the dataset only contains very regular inflections. In the original shared task data split, The overlap between the lemmata in the dev and test sets and those in the training set is 100\%. The number of paradigms which we can obtain by combining the training, dev and test sets of this dataset is around 100 for Akan, Ga and Swahili, 227 for Nyanja, 57 for Lingala and only 26 for Southern Sotho.  

For our ``wug test'', we divide the inflection tables reconstructed from this dataset into a 7:1:2 train-dev-test split, i.e. we use the same ratio as the shared task, but the division is by inflection tables rather than lemma-tag-form triples, to ensure that the lemmata used for validation and test are disjoint from those for training. 
We provide details on the data statistics in Appendix \ref{apdx:data} for reference.


\section{Experiments}
\label{sec:exp}

\paragraph{Inflection model} The Transformer \citep{vaswani2017attention} is the seq2seq architecture which produces the current state-of-the-art result on the morphological inflection task \citep{wu2020applying,vylomova-etal-2020-sigmorphon,liu-hulden-2020-analogy}. It takes the lemma and target tag(s) as input and predicts the target form character by character. Our experiments use the Transformer model implemented in Fairseq \cite{ott-etal-2019-fairseq} and adopt the same hyperparameter settings as \citet{liu-hulden-2020-analogy}. 

\smallbreak
\noindent \textbf{Evaluation metric} \ \ \ The evaluation metric is accuracy. 
For the original shared task data and experiments on 2018-languages, we train five inflection models each with different random initialization and report the average accuracy with standard deviation. Due to data scarcity, for Niger-Congo languages at the ``wug test''-like setting, we perform a 5-fold cross-validation and report the average accuracy and the standard deviation.

\smallbreak
\noindent \textbf{Common-practice test and ``wug test''} \ \ \ We first compare the performance of the Transformer in the common-practice setting and the ``wug test''-like setting. The ``common practice'' is represented by previous years' shared tasks and related work \citep{cotterell-etal-2016-sigmorphon,cotterell-etal-2017-conll,cotterell-etal-2018-conll,mccarthy-etal-2019-sigmorphon,vylomova-etal-2020-sigmorphon}; here the training data usually covers a fair number of lemmata and there is overlap between lemmata in the training and test sets. We use the shared task data to represent the common-practice setting. In the ``wug test'' setting, the lemmata to be inflected are always previously unseen. To our surprise, the performance of the Transformer at the ``wug test''-like setting is very poor despite the large amount of training triples for 2018-languages or the very regular and straightforward inflection for Niger-Congo languages. The performance is dramatically inferior to the common-practice setting, even when the number of training triples is seven times larger for Finnish, Spanish and Turkish. 

We hypothesize four reasons for the poor performance of the model under the ``wug test''-like circumstance: (1) missing copy bias from entire stem, i.e. the model can't copy a stem $abcde$ if that exact stem has never been seen during training, (2) missing copy bias on individual letters, i.e. the model can't copy letter $a$ if the letter is underrepresented in training, (3) missing copy bias on subsequences of letters, i.e. the model can't copy sequence $ab$ if the sequence is underrepresented in training, (4) some combination of all factors. To test these hypotheses, we conduct five experiments designed to help the model learn to copy with different biases by adding to the training set for each language 2,000\footnote{The choice of 2,000 is in order to match the augmentation size of \textit{+copy-dev-test-lemmas} method for 2018-languages. We did not try to tune for the best data augmentation size. Appendix \ref{apdx:aug-size} provides plots of data augmentation size comparison, where we found no consistent difference in all the languages.} dummy datapoints generated in five different ways, explained below.

\smallbreak
\noindent \textbf{\textit{+copy-dev-test-lemmas}} \ \ \ In order to test the first hypothesis that the model does not learn to copy parts of a stem it has not seen at the training stage, we augment the training data for each language by adding the lemmata in its development and test sets with a special tag \texttt{COPY}. In other words, 2000 \texttt{\small{(LEMMA, COPY, LEMMA)}} triples are added to the initial ``wug test'' training set for each language.

\begin{figure}
    \centering
    \includegraphics[width=0.9\columnwidth]{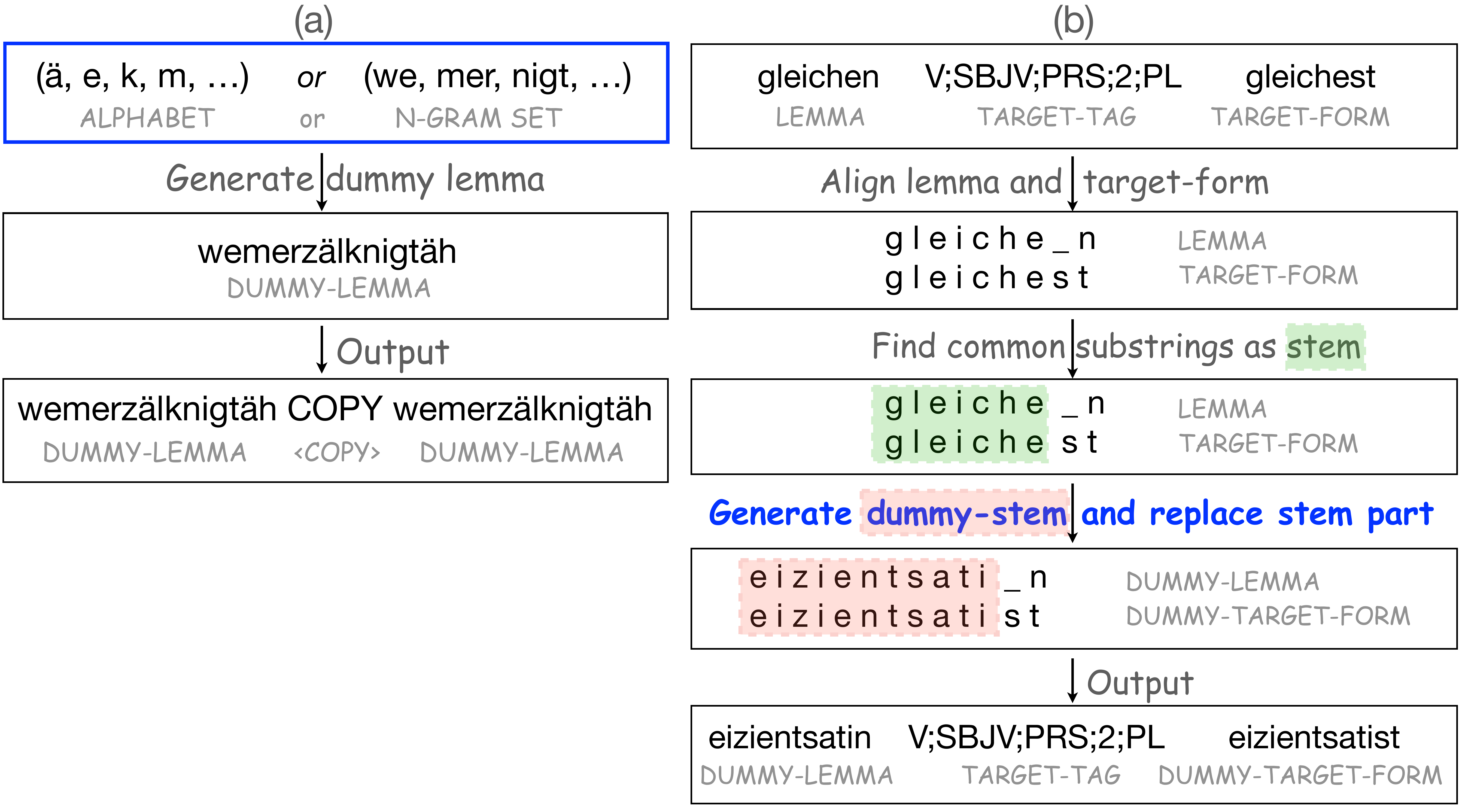}
    \caption{(a) Dummy lemma generation with a German example. \textit{+copy-2k-char} generates random strings by uniformly sampling from the alphabet, while \textit{+copy-2k-substr} samples from the set of 2-, 3- and 4-grams; (b) Data hallucination with a German example. \textit{+hall-2k-substr} is different from \textit{+hall-2k-char} in how the dummy-stem is generated.}
    \label{fig:data-aug}
\end{figure}

\smallbreak
\noindent \textbf{\textit{+copy-2k-char} and \textit{+copy-2k-substr}} \ \ Previous work found that adding random strings can help seq2seq models learn a copy bias and thus improve the performance when the training data is limited \citep{kann-schutze-2017-unlabeled}. We adopt a similar method to augment the training data with dummy lemmata generated by the process shown in Figure \ref{fig:data-aug} (a). The \textit{+copy-2k-char} method takes as input the alphabet created by collecting characters in the language's training set. Considering that a natural linguistic sub-unit of a word is a syllable, we propose to use substrings of syllable-like length for the \textit{+copy-2k-substr} method. The input of this method is the set of bigrams, trigrams and four-grams from the language's training data. For both methods, we generate the dummy lemma by uniformly sampling from the input and concatenating the sampled items to a random length between the minimum and maximum word length we see in the training data. The output of the dummy lemma generation process is a triple of a dummy lemma, a special symbol \texttt{COPY} and the dummy lemma, which is added to the initial ``wug test'' training set for data augmentation.
 


\begin{figure*}[!htb]
    \centering
    \includegraphics[width=.75\linewidth]{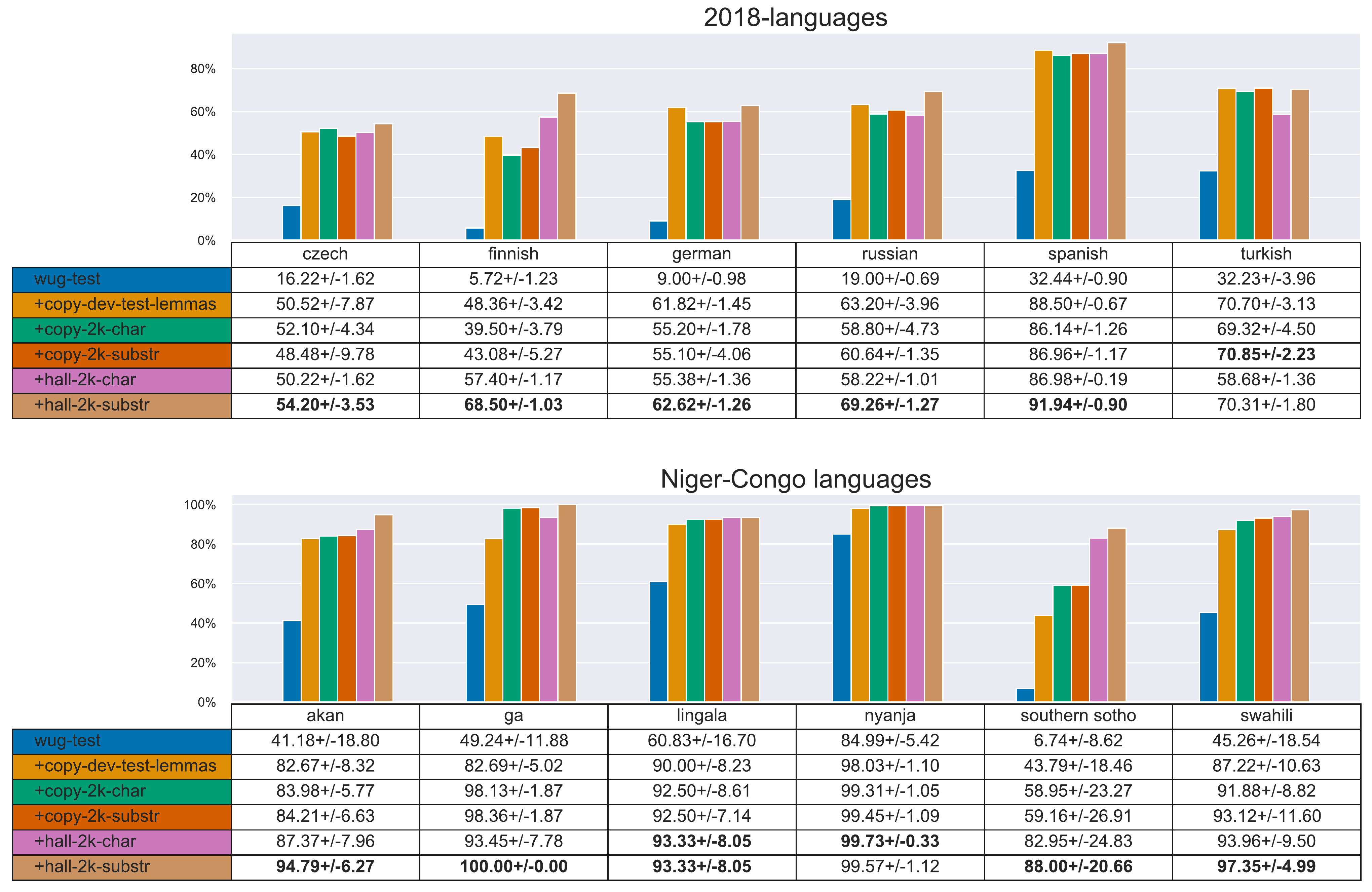}
    \caption{``Wug test'' results. \textit{+copy-2k-char} adds random strings generated with the alphabet. \textit{+copy-2k-substr} adds random strings generated with the n-gram set. \textit{+hall-2k-char} adds data hallucinated with the method by \citet{anastasopoulos-neubig-2019-pushing}. \textit{+hall-2k-substr} adds data hallucinated with our method.}
    \label{fig:wug-results}
\end{figure*}

\smallbreak
\noindent \textbf{\textit{+hall-2k-char} and \textit{+hall-2k-substr}} \ \ \ The dummy lemma generation methods do not leverage knowledge about word structure which can be inferred from the training data. \citet{silfverberg-etal-2017-data} found that augmenting training data in low-resource situations with data hallucination by replacing a hypothesized stem of the training triples with a random string is very effective. \citet{anastasopoulos-neubig-2019-pushing} improves this data hallucination method by taking into consideration of discontinuous stems as well, which is the best data hallucination method so far. We conduct the \textit{+hall-2k-char} experiment by augmenting the initial ``wug test'' training set with dummy data generated with \citet{anastasopoulos-neubig-2019-pushing}'s method. The implementation from SIGMORPHON 2020 shared task 0 baseline is used. In addition, we propose to generate the dummy stem by uniformly sampling from substrings of syllable-like length, i.e. the bigram, trigram and four-gram set. This experiment is referred to as \textit{+hall-2k-substr}. Specifically, both data hallucination methods (illustrated in Figure \ref{fig:data-aug} (b)) take as input a triple from the training set, align the lemma and the target form with the alignment method from SIGMORPHON 2016 shared task baseline \citep{cotterell-etal-2016-sigmorphon}, find the common substrings between the lemma and the target form as the stem, replace the stem with a dummy stem, and output a dummy triple 
which is adopted for data augmentation. Our proposed method is different from \citet{anastasopoulos-neubig-2019-pushing}'s method at the dummy stem generation step in two main aspects: (1) Instead of sampling from the alphabet, we sample from the set of bigrams, trigrams and four-grams. (2) Instead of forcing the dummy stem to be of the same length as the stem to be replaced, we only constrain the minimum and maximum length of the stem based on the training data. In addition, for discontinuous stems, we only replace the first part of the stem.{\footnote{Using the first part only is for implementation simplicity in the current work. It should be adjusted for languages with a large number of discontinuous stems.}}




\section{Results and discussion}
\label{sec:results}

\paragraph{``Wug test'' with data augmentation}

Figure \ref{fig:wug-results} shows results for the ``wug test''-like setting and results after augmenting the initial training set with different methods. Every language gets significant improvement with data augmentation, indicating that the Transformer model at the vanilla ``wug test'' circumstance will not learn a copy bias well. 

The substring-based data hallucination we propose, \textit{+hall-2k-substr}, achieves accuracies which are significantly higher than other methods for most languages. For Turkish and Nyanja, \textit{+hall-2k-substr} is lower than the best performance, but the difference is not significant. For Lingala, \textit{+hall-2k-substr} has the same best performance as \textit{+hall-2k-char}. The outstanding advantage of \textit{+hall-2k-substr} indicates that substrings of syllable-like length is more helpful than individual characters for data hallucination. It also provides support to the fourth hypothesis we made in section \ref{sec:exp} that the poor performance of the Transformer in the vanilla ``wug test''-like setting is due to a combination of factors including missing copying bias for letters, subsequences of letters and even entire stems.

\smallbreak
\noindent\textbf{Common practice vs ``wug test''} \ \ \ Figure \ref{fig:shared_tas_vs_full_table_vs_add_hall} plots the Transformer accuracies with standard deviations in the common-practice setting, vanilla ``wug test''-like setting, and ``wug test''-like setting with data augmentation by the substring-based data hallucination methods (\textit{+hall-2k-substr}). Though data augmentation can improve the model's performance for ``wug test'', results are still inferior to the common practice setting without any data augmentation for most languages, especially the morphologically challenging 2018-languages.

\section{Conclusion}

In this work, we examine keeping training lemmata disjoint from the evaluation sets in morphological inflection. By comparing the performance of the Transformer under the ``wug test''-like circumstances with the common practice, we find that the common-practice setting where there is overlap of lemmata has obscured the difficulty of the inflection task. We propose to augment the training data with substring-based data hallucination, and achieve significant improvement over previous data hallucination methods.

\bibliographystyle{acl_natbib}
\bibliography{acl2021}

\begin{thebibliography}{14}
\expandafter\ifx\csname natexlab\endcsname\relax\def\natexlab#1{#1}\fi

\bibitem[{Anastasopoulos and Neubig(2019)}]{anastasopoulos-neubig-2019-pushing}
Antonios Anastasopoulos and Graham Neubig. 2019.
\newblock \href {https://doi.org/10.18653/v1/D19-1091} {Pushing the limits of
  low-resource morphological inflection}.
\newblock In \emph{Proceedings of the 2019 Conference on Empirical Methods in
  Natural Language Processing and the 9th International Joint Conference on
  Natural Language Processing (EMNLP-IJCNLP)}, pages 984--996, Hong Kong,
  China. Association for Computational Linguistics.

\bibitem[{Berko(1958)}]{berko1958}
Jean Berko. 1958.
\newblock The child's learning of {E}nglish morphology.
\newblock \emph{Word}, 14(2-3):150--177.

\bibitem[{Cotterell et~al.(2018)Cotterell, Kirov, Sylak-Glassman, Walther,
  Vylomova, McCarthy, Kann, Mielke, Nicolai, Silfverberg, Yarowsky, Eisner, and
  Hulden}]{cotterell-etal-2018-conll}
Ryan Cotterell, Christo Kirov, John Sylak-Glassman, G{\'e}raldine Walther,
  Ekaterina Vylomova, Arya~D. McCarthy, Katharina Kann, Sabrina~J. Mielke,
  Garrett Nicolai, Miikka Silfverberg, David Yarowsky, Jason Eisner, and Mans
  Hulden. 2018.
\newblock \href {https://doi.org/10.18653/v1/K18-3001} {The
  {C}o{NLL}{--}{SIGMORPHON} 2018 shared task: Universal morphological
  reinflection}.
\newblock In \emph{Proceedings of the {C}o{NLL}{--}{SIGMORPHON} 2018 Shared
  Task: Universal Morphological Reinflection}, pages 1--27, Brussels.
  Association for Computational Linguistics.

\bibitem[{Cotterell et~al.(2017)Cotterell, Kirov, Sylak-Glassman, Walther,
  Vylomova, Xia, Faruqui, K{\"u}bler, Yarowsky, Eisner, and
  Hulden}]{cotterell-etal-2017-conll}
Ryan Cotterell, Christo Kirov, John Sylak-Glassman, G{\'e}raldine Walther,
  Ekaterina Vylomova, Patrick Xia, Manaal Faruqui, Sandra K{\"u}bler, David
  Yarowsky, Jason Eisner, and Mans Hulden. 2017.
\newblock \href {https://doi.org/10.18653/v1/K17-2001} {{C}o{NLL}-{SIGMORPHON}
  2017 shared task: Universal morphological reinflection in 52 languages}.
\newblock In \emph{Proceedings of the {C}o{NLL} {SIGMORPHON} 2017 Shared Task:
  Universal Morphological Reinflection}, pages 1--30, Vancouver. Association
  for Computational Linguistics.

\bibitem[{Cotterell et~al.(2016)Cotterell, Kirov, Sylak-Glassman, Yarowsky,
  Eisner, and Hulden}]{cotterell-etal-2016-sigmorphon}
Ryan Cotterell, Christo Kirov, John Sylak-Glassman, David Yarowsky, Jason
  Eisner, and Mans Hulden. 2016.
\newblock \href {https://doi.org/10.18653/v1/W16-2002} {The {SIGMORPHON} 2016
  shared {T}ask{---}{M}orphological reinflection}.
\newblock In \emph{Proceedings of the 14th {SIGMORPHON} Workshop on
  Computational Research in Phonetics, Phonology, and Morphology}, pages
  10--22, Berlin, Germany. Association for Computational Linguistics.

\bibitem[{Kann and Sch{\"u}tze(2017)}]{kann-schutze-2017-unlabeled}
Katharina Kann and Hinrich Sch{\"u}tze. 2017.
\newblock \href {https://doi.org/10.18653/v1/W17-4111} {Unlabeled data for
  morphological generation with character-based sequence-to-sequence models}.
\newblock In \emph{Proceedings of the First Workshop on Subword and Character
  Level Models in {NLP}}, pages 76--81, Copenhagen, Denmark. Association for
  Computational Linguistics.

\bibitem[{Kirov et~al.(2018)Kirov, Cotterell, Sylak-Glassman, Walther,
  Vylomova, Xia, Faruqui, Mielke, McCarthy, K{\"u}bler, Yarowsky, Eisner, and
  Hulden}]{kirov-etal-2018-unimorph}
Christo Kirov, Ryan Cotterell, John Sylak-Glassman, G{\'e}raldine Walther,
  Ekaterina Vylomova, Patrick Xia, Manaal Faruqui, Sabrina~J. Mielke, Arya
  McCarthy, Sandra K{\"u}bler, David Yarowsky, Jason Eisner, and Mans Hulden.
  2018.
\newblock \href {https://www.aclweb.org/anthology/L18-1293} {{U}ni{M}orph 2.0:
  {U}niversal {M}orphology}.
\newblock In \emph{Proceedings of the Eleventh International Conference on
  Language Resources and Evaluation ({LREC} 2018)}, Miyazaki, Japan. European
  Language Resources Association (ELRA).

\bibitem[{Liu and Hulden(2020)}]{liu-hulden-2020-analogy}
Ling Liu and Mans Hulden. 2020.
\newblock \href {https://www.aclweb.org/anthology/2020.coling-main.257}
  {Analogy models for neural word inflection}.
\newblock In \emph{Proceedings of the 28th International Conference on
  Computational Linguistics}, pages 2861--2878, Barcelona, Spain (Online).
  International Committee on Computational Linguistics.

\bibitem[{McCarthy et~al.(2019)McCarthy, Vylomova, Wu, Malaviya, Wolf-Sonkin,
  Nicolai, Kirov, Silfverberg, Mielke, Heinz, Cotterell, and
  Hulden}]{mccarthy-etal-2019-sigmorphon}
Arya~D. McCarthy, Ekaterina Vylomova, Shijie Wu, Chaitanya Malaviya, Lawrence
  Wolf-Sonkin, Garrett Nicolai, Christo Kirov, Miikka Silfverberg, Sabrina~J.
  Mielke, Jeffrey Heinz, Ryan Cotterell, and Mans Hulden. 2019.
\newblock \href {https://doi.org/10.18653/v1/W19-4226} {The {SIGMORPHON} 2019
  shared task: Morphological analysis in context and cross-lingual transfer for
  inflection}.
\newblock In \emph{Proceedings of the 16th Workshop on Computational Research
  in Phonetics, Phonology, and Morphology}, pages 229--244, Florence, Italy.
  Association for Computational Linguistics.

\bibitem[{Ott et~al.(2019)Ott, Edunov, Baevski, Fan, Gross, Ng, Grangier, and
  Auli}]{ott-etal-2019-fairseq}
Myle Ott, Sergey Edunov, Alexei Baevski, Angela Fan, Sam Gross, Nathan Ng,
  David Grangier, and Michael Auli. 2019.
\newblock \href {https://doi.org/10.18653/v1/N19-4009} {fairseq: A fast,
  extensible toolkit for sequence modeling}.
\newblock In \emph{Proceedings of the 2019 Conference of the North {A}merican
  Chapter of the Association for Computational Linguistics (Demonstrations)},
  pages 48--53, Minneapolis, Minnesota. Association for Computational
  Linguistics.

\bibitem[{Silfverberg et~al.(2017)Silfverberg, Wiemerslage, Liu, and
  Mao}]{silfverberg-etal-2017-data}
Miikka Silfverberg, Adam Wiemerslage, Ling Liu, and Lingshuang~Jack Mao. 2017.
\newblock \href {https://doi.org/10.18653/v1/K17-2010} {Data augmentation for
  morphological reinflection}.
\newblock In \emph{Proceedings of the {C}o{NLL} {SIGMORPHON} 2017 Shared Task:
  Universal Morphological Reinflection}, pages 90--99, Vancouver. Association
  for Computational Linguistics.

\bibitem[{Vaswani et~al.(2017)Vaswani, Shazeer, Parmar, Uszkoreit, Jones,
  Gomez, Kaiser, and Polosukhin}]{vaswani2017attention}
Ashish Vaswani, Noam Shazeer, Niki Parmar, Jakob Uszkoreit, Llion Jones,
  Aidan~N Gomez, Lukasz Kaiser, and Illia Polosukhin. 2017.
\newblock Attention is all you need.
\newblock \emph{arXiv preprint arXiv:1706.03762}.

\bibitem[{Vylomova et~al.(2020)Vylomova, White, Salesky, Mielke, Wu, Ponti,
  Hall~Maudslay, Zmigrod, Valvoda, Toldova, Tyers, Klyachko, Yegorov,
  Krizhanovsky, Czarnowska, Nikkarinen, Krizhanovsky, Pimentel,
  Torroba~Hennigen, Kirov, Nicolai, Williams, Anastasopoulos, Cruz, Chodroff,
  Cotterell, Silfverberg, and Hulden}]{vylomova-etal-2020-sigmorphon}
Ekaterina Vylomova, Jennifer White, Elizabeth Salesky, Sabrina~J. Mielke,
  Shijie Wu, Edoardo~Maria Ponti, Rowan Hall~Maudslay, Ran Zmigrod, Josef
  Valvoda, Svetlana Toldova, Francis Tyers, Elena Klyachko, Ilya Yegorov,
  Natalia Krizhanovsky, Paula Czarnowska, Irene Nikkarinen, Andrew
  Krizhanovsky, Tiago Pimentel, Lucas Torroba~Hennigen, Christo Kirov, Garrett
  Nicolai, Adina Williams, Antonios Anastasopoulos, Hilaria Cruz, Eleanor
  Chodroff, Ryan Cotterell, Miikka Silfverberg, and Mans Hulden. 2020.
\newblock \href {https://doi.org/10.18653/v1/2020.sigmorphon-1.1} {{SIGMORPHON}
  2020 shared task 0: Typologically diverse morphological inflection}.
\newblock In \emph{Proceedings of the 17th SIGMORPHON Workshop on Computational
  Research in Phonetics, Phonology, and Morphology}, pages 1--39, Online.
  Association for Computational Linguistics.

\bibitem[{Wu et~al.(2020)Wu, Cotterell, and Hulden}]{wu2020applying}
Shijie Wu, Ryan Cotterell, and Mans Hulden. 2020.
\newblock Applying the transformer to character-level transduction.
\newblock \emph{arXiv preprint arXiv:2005.10213}.

\end{thebibliography}

\clearpage
\pagebreak

\appendix

\section{Data information}
\label{apdx:data}

\begin{table}[!htb]
    \centering
    \begin{adjustbox}{width=\linewidth}
    \begin{tabular}{cccc|ccc|cc}
    \toprule
    & \multicolumn{3}{c|}{triple-counts} & \multicolumn{3}{c|}{lemma-counts} & \multicolumn{2}{c}{lemma-overlap (\%)} \\
    Language & train & dev & test & train & dev & test & dev-train & test-train \\
    \hline
    czech & 1000 & 1000 & 1000 & 848 & 848 & 849 & 24.53 & 20.38 \\
    finnish & 1000 & 1000 & 1000 & 985 & 983 & 987 & 2.34 & 3.04 \\
    german & 1000 & 1000 & 1000 & 961 & 945 & 962 & 9.42 & 9.46 \\
    russian & 1000 & 1000 & 1000 & 973 & 985 & 977 & 3.65 & 3.79 \\
    spanish & 1000 & 1000 & 1000 & 906 & 902 & 922 & 15.74 & 16.49 \\
    turkish & 906 & 928 & 912 & 764 & 802 & 779 & 26.06 & 26.57 \\
    \bottomrule
    \end{tabular}
    \end{adjustbox}
    \caption{CoNLL-SIGMORPHON 2018 shared task 1 medium-size data information.}
    \label{tab:2018-med-data}
\end{table}

\begin{table}[!htb]
    \centering
    \begin{adjustbox}{width=0.85\linewidth}
    \begin{tabular}{cc|c|cc}
    \toprule
    & triple-counts & lemma-counts & \multicolumn{2}{|c}{lemma-overlap (\%)} \\
    Language & train & train & dev-train & test-train \\
    \hline
    czech & 1582 & 100 & 0 & 0 \\
    finnish & 7136 & 100 & 0 & 0 \\
    german & 1290 & 100 & 0 & 0 \\
    russian & 1311 & 100 & 0 & 0 \\
    spanish & 7132 & 100 & 0 & 0 \\
    turkish & 7632 & 100 & 0 & 0 \\

    \bottomrule
    \end{tabular}
    \end{adjustbox}
    \caption{Data information of the training set we create for 2018-languages. We use the same dev and test sets as CoNLL-SIGMORPHON 2018 shared task 1.}
    \label{tab:our-data}
\end{table}

\begin{table}[!htb]
    \centering
    \begin{adjustbox}{width=\linewidth}
    \begin{tabular}{lccc|ccc|cc}
    \toprule
    & \multicolumn{3}{c|}{triple-counts} & \multicolumn{3}{c|}{lemma-counts} & \multicolumn{2}{c}{lemma-overlap (\%)} \\
    Language & train & dev & test & train & dev & test & dev-train & test-train \\
    \hline
    akan & 2793 & 380 & 763 & 96 & 94 & 95 & 100.0 & 100.0 \\
    ga & 607 & 79 & 169 & 95 & 59 & 80 & 100.0 & 100.0 \\
    lingala & 159 & 23 & 46 & 57 & 23 & 34 & 100.0 & 100.0 \\
    nyanja & 3031 & 429 & 853 & 227 & 199 & 226 & 100.0 & 100.0 \\
    southern sotho & 345 & 50 & 99 & 26 & 24 & 25 & 100.0 & 100.0 \\
    swahili & 3374 & 469 & 910 & 97 & 97 & 96 & 100.0 & 100.0 \\
    \bottomrule
    \end{tabular}
    \end{adjustbox}
    \caption{Data information of Niger-Congo languages from SIGMORPHON 2020 shared task 0.}
    \label{tab:niger-congo-data}
\end{table}

\newpage
\section{Data augmentation size comparison}
\label{apdx:aug-size}

\begin{figure}[!htb]
    \centering
    \includegraphics[width=\linewidth]{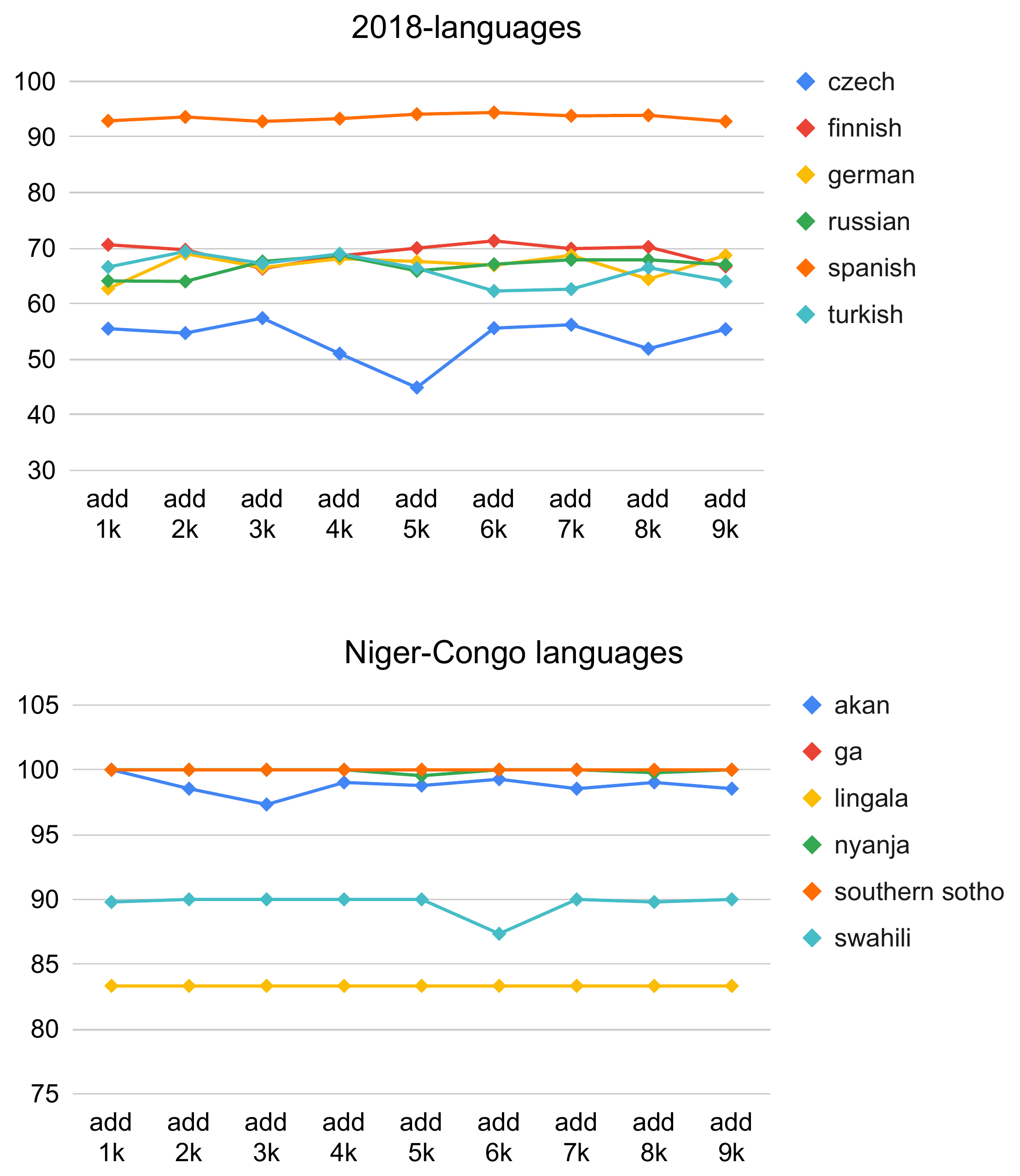}
    \caption{Performance on the dev set in ``wug test'' after \textbf{adding different amounts of dummy data} generated with our substring-based hallucination method.}
    \label{fig:augsize}
\end{figure}



\end{document}